# Time Series Forecasting (TSF) Using Various Deep Learning Models

Jimeng Shi, Mahek Jain, Giri Narasimhan

*Abstract*—Time Series Forecasting (TSF) is used to predict the target variables at a future time point based on the learning from previous time points. To keep the problem tractable, learning methods use data from a fixed length window in the past as an explicit input. In this paper, we study how the performance of predictive models change as a function of different look-back window sizes and different amounts of time to predict into the future. We also consider the performance of the recent attention-based Transformer models, which has had good success in the image processing and natural language processing domains. In all, we compare four different deep learning methods (RNN, LSTM, GRU, and Transformer) along with a baseline method. The dataset (hourly) we used is the Beijing Air Quality Dataset from the UCI *website*, which includes a multivariate time series of many factors measured on an hourly basis for a period of 5 years (2010-14). For each model, we also report on the relationship between the performance and the look-back window sizes and the number of predicted time points into the future. Our experiments suggest that Transformer models have the best performance with the lowest Mean Average Errors (MAE = 14.599, 23.273) and Root Mean Square Errors (RSME = 23.573, 38.131) for most of our single-step and multi-steps predictions. The best size for the look-back window to predict 1 hour into the future appears to be one day, while 2 or 4 days perform the best to predict 3 hours into the future.

## I. INTRODUCTION

TIME series is a sequence of repeated observations of a given set of $m$ variables over a period time [1]. Examples include stock prices, precipitation, volume of traffic in a communication or transportation network, and much more. It can be mathematically defined as $\{x_1, x_2, ..., x_T\}$, where $t = 1, 2, ..., T$ represents the elapsed time, and $x_t$ measured at time $t$, denotes a vector of $m$ random variables [2]. Time Series Forecasting (TSF) is used to predict the distribution of the target variables at a future time based on the past observations of the time series. Many useful temporal inferencing problems have been considered including *filtering, smoothing, predictions* of unobserved past events or alternative histories [3]. Efficient and accurate predictions of the future have significant applications in many different domains – finance [4], traffic [5], engineering [6], healthcare [7], weather [8], and more. Better predictions can lead to better investments, better management of traffic anomalies, improved management of resources during supply chain fluctuations, efficient handling of spikes in infections during epidemics, and much more, smarter decisions in agriculture and environment to handle foreseeable weather patterns or environmental disasters. In this paper, we consider the problem of predicting air quality (PM2.5) to better manage urban air pollution, which has become a serious threat to the environmental and human health with the acceleration of industrialization [9].

The models used to capture time series can be divided into 3 categories: traditional models, machine learning models, and deep learning models. Traditional models can be divided into linear and non-linear ones [1]. Autoregressive Moving Average (ARMA) [10, 11] and Autoregressive Integrated Moving Average (ARIMA) are two well-known linear models, which can solve *stationary* and *non-stationary* time series respectively. A time series is stationary if its mean and variance are constant (time-independent; no *drift*) for any period. ARIMA is used to model non-stationary time series by first transforming it to make it stationary. Variants of ARIMA include Autoregressive Fractionally Integrated Moving Average (ARFIMA) [12] and Seasonal Autoregressive Integrated Moving Average (SARIMA) [10, 13]. As with ARIMA, SARIMA first transforms the time series to make it stationary by eliminating the seasonal component. Among the non-linear models, Autoregressive Conditional Heteroskedasticity (ARCH) [14, 15, 16] and its variants like Generalized ARCH (GARCH) [14, 15, 16], Exponential Generalized ARCH (EGARCH) [16], Threshold Autoregressive (TAR) [17], Non-linear Autoregressive (NAR) [18, 19], and Non-linear Moving Average (NMA) [20, 21]. The primary limitations of the traditional TSF models are that they apply regression to a fixed set of factors from the most recent historical data to generate the predictions. Second, traditional methods are iterative and are often sensitive to how the process is seeded. Third, stationarity is a strict condition, and it is difficult to achieve stationarity of volatile time series by merely addressing drift, seasonality, autocorrelation, and heteroskedasticity. Hence the need for machine learning models.

Standard machine learning models such as Support Vector Machines (SVM) [22, 23] have been used in this context, as have hybrid approaches combining ARIMA with SVM [24] and Neural Networks [25]. Artificial Neural Network (ANN) [26, 27] and deep learning NNs [28] have been shown to have better

Jimeng Shi is with the Knight Foundation School of Computing and Information Sciences, Florida International University, Miami, FL 33199 USA (e-mail: jshi008@fiu.edu).

Mahek Jain is with the Department of Computer Science and Engineering, Rashtreeya Vidyalaya College of Engineering (RVCE), Bangalore 560059 India (e-mail: mhkjain25@gmail.com).

Giri Narasimhan is with the Knight Foundation School of Computing and Information Sciences, Florida International University, Miami, FL 33199 USA (corresponding author, phone: 305-348-3748; e-mail: giri@fiu.edu).

performance than the traditional approaches. Machine learning approaches best suited for time series forecasting include Recurrent Neural Network (RNN) [29], Long Short-term Memory (LSTM) [30], and GRUs. Improved forecasting has been achieved by using attention-based methods called *Transformers* [31]. A review of the literature on deep learning for TSF can be found here [32, 33]. Transformers, introduced in 2018 [34], has been successfully used in many domains, including speech recognition [35], machine translation [34], and computer vision [36, 37, 38].

While there is considerable prior work on modeling time series, there is a lack of knowledge on the practical considerations in using deep learning for time series forecasting. The problem of how much of the past (size of look-back window) or the how far into the future we can reliably predict has not been investigated. The aims of this work are: (1) to apply and validate deep learning models (RNN, LSTM, GRU, Transformer) for time series forecasting and compare their corresponding performance; (2) to assess the strengths and weaknesses of these models; and (3) to understand the impact of the size of look-back window and the length of time of future predictions on the prediction accuracy.

For the Beijing Air Quality Dataset from the UCI website, which includes a multivariate time series of many factors measured on an hourly basis for a period of 5 years (2010-14), our experiments show that the Transformer models have the best performance for our predictions. We also pinpoint the best look-back window sizes to use for the best predictions at a specified future time.

## II. METHODOLOGY

Using historical data, deep learning models learn a functional relationship between input features and future values of the target variable. The resulting model can provide predictions for the target variable at future time points. Given a time series, $\{x_1, x_2, ..., x_T\}$, where $x_t$ is a vector of $m$ input features observed or measured at time $t$, the task is to develop a model to predict a target variable $y_{t+k}$ at a future time point, $t + k$, using historical data, i.e., the time series ending at time $t - 1$, $\{..., x_{t-2}, x_{t-1}\}$. To make the input to the model of a uniform length, we used a fixed length sliding time window of size $w$, as shown in Fig. 1. Data was transformed by min-max scaling using Eq. (5). Mathematically, the functional relationship learned by the machine learning models can be written down as shown in Eq. (1):

$$\hat{y}_{t+k} = f_k(x_{t-w}, ..., x_{t-1}, y_{t-w}, ..., y_{t-1}) \quad (1)$$

where $\hat{y}_{t+k}$ is the target variable forecast for time $t + k$; $k$ is the length of the time into the future for which the target variable is to be predicted; $y_{t-w}, ..., y_{t-1}$ are the target values observed from time $t - w$ to $t - 1$; $x_{t-w}, ..., x_{t-1}$ are the vector of $m$ observed input features from time $t - w$ to $t - 1$; $f_k$ is the function learned by deep learning models; $m$ is the number of input features; $w$ is the size of the window used as input. A detailed example of the setup is provided in the *Appendix*.

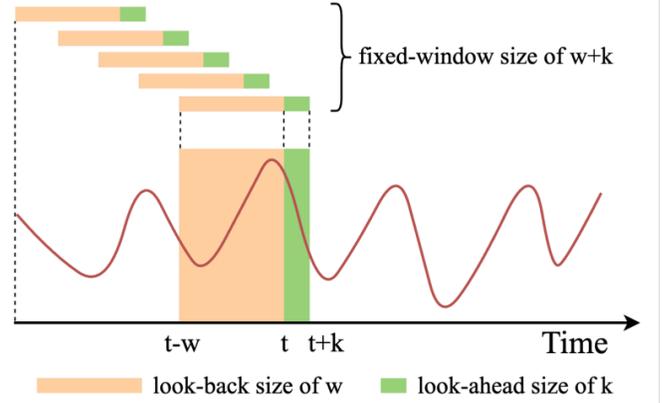

Fig. 1 Using a sliding window of width $w$ to construct the training set for predicting $k$ time steps in the future.

## III. DEEP LEARNING FRAMEWORKS

This section will provide brief descriptions of the deep learning models used in this work: Recurrent Neural Networks (RNN), Long Short-term Memory (LSTM), Gated Recurrent Units (GRU) and Transformers.

### A. Recurrent Neural Networks (RNN)

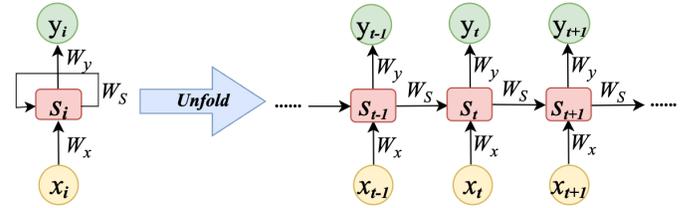

Fig. 2 Structure of the RNN

RNNs are best suited for modeling time series data [39]. RNNs use neural networks to model the functional relationship between input features in the recent past to a target variable in the future. As shown in Fig. 2, an RNN recurrently learns from a training set of historical data by focusing on the transitions of an internal (hidden) state from time $t - 1$ to time $t$. The resulting model is determined by three parameter matrices, $W_x$, $W_y$, and $W_s$, and two bias vectors, $b_s$ and $b_y$, that help define the model. The output $y_t$ depends on the internal state, $S_t$, which depends on both current input $x_t$ and the previous state $S_{t-1}$ [40]. The computational process of each hidden state (hidden unit or hidden cell) is described in Fig. 3. Mathematically, it is given as follows:

$$\begin{aligned} S_t &= tanh(W_{xs} \cdot (x_t \oplus S_{t-1}) + b_s) \text{ and} \\ y_t &= \sigma(W_y \cdot S_t + b_y), \end{aligned} \quad (2)$$

where $x_t \in \mathbb{R}^m$ is the input vector of $m$ input features at time $t$; $W_{xs} \in \mathbb{R}^{n \times (m+n)}$ and $W_y \in \mathbb{R}^{n \times n}$ are parameter matrices; $n$ is the number of neurons in the RNN layer; $b_s \in \mathbb{R}^N$ and $b_y \in \mathbb{R}^N$ are bias vectors for the internal state and output, respectively; $\sigma$ is the sigmoid activation function; $S_t$ is the internal (hidden) state; and $x_t \oplus S_{t-1}$ is the concatenation of vectors, $x_t$ and $S_{t-1}$.

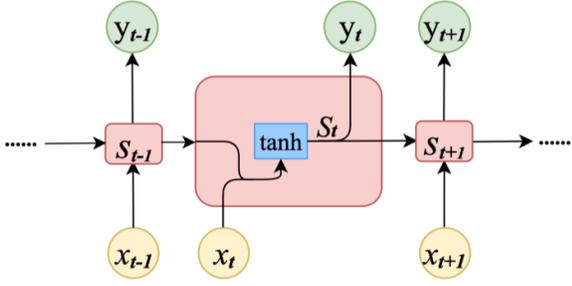

Fig. 3 Computation process describing the RNN

The biggest drawback of RNNs is that due to repeated multiplication of the recurrent weight matrix they suffer from the gradient vanishing problem [41, 42, 43], because of which the gradient becomes too small over time and the RNN ends up remembering information only for small durations of time.

*B. Long Short-term Model (LSTM)*

Long Short-term Memory (LSTM) networks are a variant of RNNs that partially address the vanishing gradient problem [44] and learn longer-term dependencies in the time series data. Additional details on LSTMs can be found in [45, 46]. They are described at time $t$ in terms of an internal (hidden) state, $S_t$, and a cell state, $C_t$. As shown in Fig. 4, $C_t$ has three different dependencies [46]: (1) previous cell state, $C_{t-1}$; (2) previous internal state, $S_{t-1}$; and (3) input at the current time point $x_t$. The process displayed in Fig. 4 allows for removal/filtering, multiplication/combining, and addition of information using forget gate, input gate, addition gate, and output gate, implementing the functions $f_t, i_t, \tilde{C}_t$, and $O_t$, respectively, thus allowing for finer control over learning longer-term dependencies.

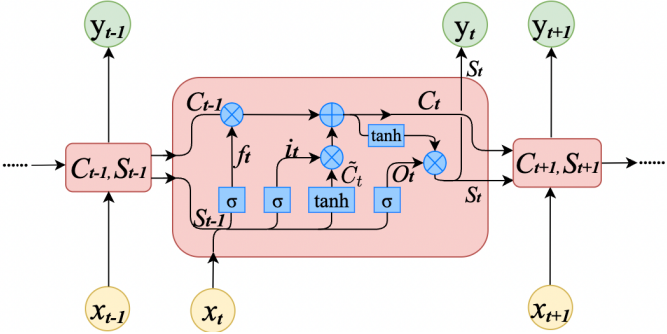

Fig. 4 Computation process involved in an LSTM

$$\begin{aligned}
f_t &= \sigma(W_f \cdot (x_t \oplus S_{t-1}) + b_f); \\
i_t &= \sigma(W_i \cdot (x_t \oplus S_{t-1}) + b_i); \\
\tilde{C}_t &= tanh(W_C \cdot (x_t \oplus S_{t-1}) + b_C); \\
C_t &= f_t \cdot C_{t-1} + i_t \cdot \tilde{C}_t; \\
O_t &= \sigma(W_O \cdot (x_t \oplus S_{t-1}) + b_O); \\
S_t &= tanh(C_t) \cdot O_t; \text{ and} \\
y_t &= \sigma(W_y \cdot S_t + b_y),
\end{aligned} \quad (3)$$

where $x_t \in \mathbb{R}^m$ is the input vector of $m$ input features at time $t$; $W_f, W_i, W_C, W_O \in \mathbb{R}^{n \times (m+n)}$ and $W_y \in \mathbb{R}^{n \times n}$ are parameter matrices; $n$ is the number of neurons in the LSTM layer; $b_f, b_i, b_C, b_O, b_y \in \mathbb{R}^n$ are bias vectors; $\sigma$ is the sigmoid activation function; and $S_t$ is the internal (hidden) state. The functions $f_t, i_t, \tilde{C}_t$, and $O_t$ are implemented by the forget gate, input gate, addition gate, and output gate, respectively.

*C. Gated Recurrent Unit (GRU)*

Gated Recurrent Units (GRU) is a variant of LSTMs to further address the vanishing gradient problem [47, 48]. As shown in Fig. 5, the novelty in this method is in the use of an update gate, a reset gate and a third gate, implementing the functions $z_t$, $r_t$ and $\tilde{S}_t$, respectively. Each gate plays a different role in controlling how to filter, use, and combine prior information The first term in the expression for the next state given by $(1 - z_t) \cdot S_{t-1}$ decides what to retain from the past while $z_t \cdot \tilde{S}_t$ determines what to collect from the current memory content.

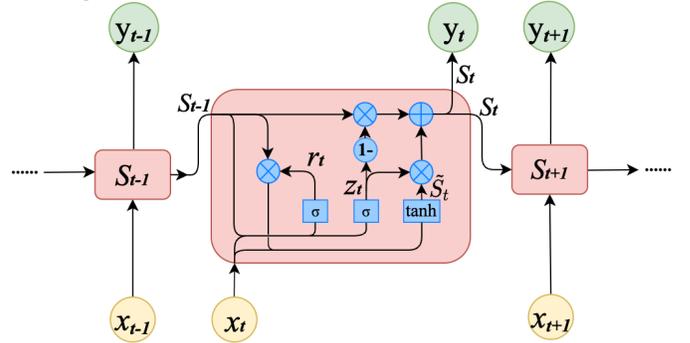

Fig. 5 Computation process of GRU

$$\begin{aligned}
r_t &= \sigma(W_r \cdot (x_t \oplus S_{t-1}) + b_r); \\
z_t &= \sigma(W_z \cdot (x_t \oplus S_{t-1}) + b_z); \\
\tilde{S}_t &= tanh(W_S \cdot (x_t \oplus S_{t-1} \cdot r_t) + b_S); \text{ and} \\
S_t &= (1 - z_t) \cdot S_{t-1} + z_t \cdot \tilde{S}_t; \text{ and} \\
y_t &= \sigma(W_y \cdot S_t + b_y),
\end{aligned} \quad (4)$$

where $x_t \in \mathbb{R}^m$ is the input vector of $m$ input features at time $t$; $W_r, W_z, W_S \in \mathbb{R}^{n \times (m+n)}$ and $W_y \in \mathbb{R}^{n \times n}$ are parameter matrices; $n$ is the number of neurons in the GRU layer; $b_r, b_z, b_S, b_y \in \mathbb{R}^n$ are bias vectors; $\sigma$ is the sigmoid activation function; and $S_t$ is the internal (hidden) state. The functions $z_t$, $r_t$ and $\tilde{S}_t$ are implemented by the update gate, reset gate, and the third gate, respectively.

*D. Transformer Model*

LSTMs and GRUs partially address the vanishing gradient problem of RNNs. However, the use of the hyperbolic tangent and the sigmoid functions as the activation function continues to cause gradient decay in the deeper layers. The transformer networks are known to have the best performance for time series because of their use of the *attention* feature, which allows them to selectively weight important information from the past [34]. Fig. 6 shows a schematic of the transformer network. It consists of an *encoder* and a *decoder* part. As before, $w$ is the size of look-back window and $k$ is the number of steps to predict in the future. In Fig. 6, the decoder part has a *Masked Attention* mechanism in the decoder, and a *Multi-Head*

*Attention* mechanism to select from the encoder output that will become the feature vector for the decoder.

Transformer is not a recurrent network, but uses *positional encoding* to mark the temporal ordering of the data. Encoder is provided the data from look-back window of size *w* and outputs a feature vector to be used by the decoder. During training, the decoder is also provided the future data that it is expected to model along with the output of the encoder. The attention feature of the transformer network helps it to learn to pay attention to important features and past trends. More details are in [34, 49, 50].

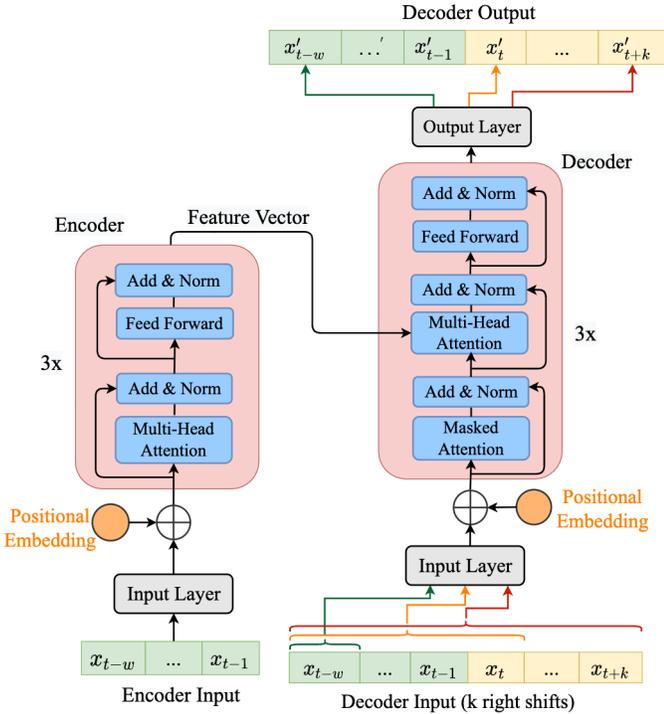

Fig. 6 Structure of the Transformer-based forecasting model

## IV. DATA AND EXPERIMENTS

We applied the four machine learning techniques to the Beijing Air Quality data set from the UCI Website [51] to perform Time Series Forecasting (TSF) for air quality predictions. Two types of experiments were performed, one referred to as a "Single step", where the next time point was predicted using data from the previous time points, and a second one referred to as a "Multi step", where the next multiple time points were predicted using data from prior time points.

### A. Dataset

The dataset we used is the hourly Beijing Air Quality Dataset from the UCI website [51], which includes data for the five-year time period from January 1, 2010 to December 31, 2014. The data was collected hourly and the data set has 43,824 rows and 13 columns. The first column is simply an index and was ignored for the analysis. The four columns labeled as year, month, day, and hour, were combined into a single feature called "year-month-day-hour". The 'PM2.5' column is the target variable. All other variables (along with time) were used as input features. The column names and explanations are specified in TABLE I. The time series for all input and target features except time and 'cbwd' are plotted in Fig. 7.

TABLE I
EXPLANATION OF VARIABLES IN THE RAW DATASET

| Comments | Variable | Explanation |
| --- | --- | --- |
| Ignored | No | Row number |
| Combined into variable 'time' | year | Year of data in this row |
| | month | Month of data in this row |
| | day | Day of data in this row |
| | hour | Hour of data in this row |
| Target | PM2.5 | PM2.5 concentration |
| Additional input features | DEWP | Dew Point |
| | TEMP | Temperature |
| | PRES | Pressure |
| | cbwd | Combined wind direction |
| | Iws | Cumulated wind speed |
| | Is | Cumulated hours of snow |
| | Ir | Cumulated hours of rain |

A small number of rows (24 out of 43,824) were discarded because of missing data. One-hot embedding was applied to the categorical feature of *wind direction* and the data were normalized values to a range of [0, 1] using the Min-Max normalization technique as shown in Eq. (5) [52, 53].

$$x' = \frac{x - x_{min}}{x_{max} - x_{min}} \quad (5)$$

The data were divided into training set (first 70% rows) and test set (last 30% rows).

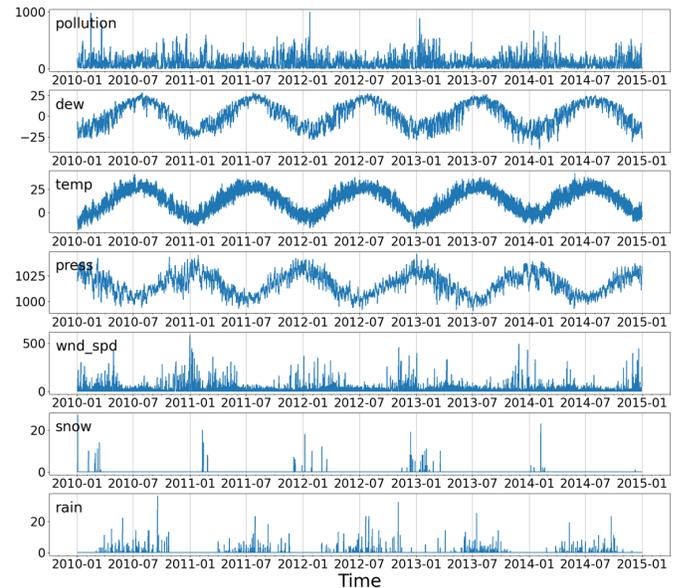

Fig. 7 Time series of the 7 input features in the dataset (wind direction time series is not shown here)

### B. Experiments

The experiments with $k = 1$ predict one time step into the future and will be referred to as *single-step* predictions, while the experiments with $k > 1$ predict one or more time points further into the future and will be referred to as *multi-step* predictions. Both sets of experiments were performed with different values of $w$, the look-back window representing the portion of the recent past used as input. Window sizes of 1, 2, 4, 8, and 16 days were used in both sets of experiments. The exponential choice of the window size was chosen to understand the impact of the window size on the prediction accuracy. The multi-step predictions were used to predict the air quality values at time points 1, 2, 4, 8, and 16 hours into the future. For each of 4 deep learning models, different hyperparameters settings were tried as shown in TABLE II. The learning rates (0.00001, 0.00005, 0.0001, 0.0005, 0.001), batch sizes (128, 256, 512) and optimizer (Adam, SGD) parameters were varied as shown, performing experiments with each possible combination.

TABLE II
SETTINGS OF VARIOUS DEEP LEARNING MODELS

|  | Epoch | LR | Batch | Optimizer |
|---|---|---|---|---|
| RNN | 100; 200 | 0.0005; 0.00001 | 256; 512 | Adam |
| LSTM | 100; 200 | 0.0005; 0.00001 | 256; 512 | Adam |
| GRU | 100; 200 | 0.0005; 0.00001 | 256; 512 | Adam |
| Transformer | 200; 300 | 0.0005; 0.00005 | 256; 512 | AdamW [54] |

*C. Measures of Evaluation*

We used the Mean Squared Error (MSE) as the loss function. The training and testing loss were computed as a function of the epochs to detect possible overfitting. Fig. 8 and Fig. 9 show the predicted and observed values of air quality for a short period of time from 2013-07-04-09:00 to 2013-07-19-08:00. The Mean Absolute Error (MAE) and Root Mean Squared Error (RMSE) were computed using the standard formulae shown below.

**Mean Absolute Error (MAE)**

$$MAE = \frac{1}{n}\sum_{i=1}^{n}|y_i - \hat{y}_i| \quad (6)$$

**Root of Mean Squared Error (RMSE)**

$$RMSE = \sqrt{\frac{1}{n}\sum_{i=1}^{n}(y_i - \hat{y}_i)^2} \quad (7)$$

## V. RESULTS

*A. Predict Multiple Timesteps Ahead*

For a fixed look-back window size (e.g., 4 Days (96 hours)), we investigate how the model performance deteriorates as we increase the value of $k$, which is the amount of time into the future for which we predict the time series value. It is safe to expect that the performance goes down as $k$ increases. This is confirmed by the fact that each column in TABLE III has MAE and RMSE values increases with $k$. As seen in TABLE III, the transformer models perform better than RNNs, LSTMs, and GRUs in 80% of the experiments. The prediction performance drops sharply when we are required to predict more than 4 hours into the future.

*B. Different Look-back Window Sizes*

Next, we investigate how the performance of the single-step and multi-step predictions are impacted by the size of the look-back window, $w$. Experiments were performed with $w = 24, 48, 96, 192$, and 384 hours.

**Single-step predictions:** TABLE IV summarizes the results of our experiments. The transformer network model outperforms the other methods for larger values of $w$ ($\geq 96\ hours$), which is consistent with the known strengths of the attention-based approach. For smaller window sizes (24 or 48 hours), GRUs and LSTMs perform better than RNNs, which is consistent with the claim that GRUs and LSTMs have longer-term memories than RNNs and have partly addressed the vanishing gradient problem. Therefore, for single-step predictions, LSTMs and GRUs are a better choice when only small window sizes can be chosen. Unfortunately, Transformer networks fail to deliver better performance even when it uses considerably larger window sizes, perhaps because of increased noise levels with larger windows. Furthermore, smaller windows are likely to lead to more efficient methods. Note that a naïve *baseline* approach for prediction that merely reports the values of the time series at the previous time point has MAE and RMSE values of 16.624 and 26.828, respectively.

**Multi-step predictions:** TABLE IV shows the results of our experiments with predicting $k = 3$ hours into the future for different values of $w$. The transformer network outperforms all the other tools. Surprisingly, the performance change is not monotonic because minimum values are reached for $w = 48$ or 96 hours, suggesting that these might be optimal values for the learning methods. Fig. 9 visualizes the predictions for our experiments; T1, T2, T3 curves represent the experimental results for $k = 1, 2$, and 3 hour.

TABLE III
PERFORMANCE (MAE AND RSME) OF MULTI-STEP PREDICTION SHOWN AS A FUNCTION OF $k$, THE NUMBER OF HOURS INTO THE FUTURE FOR WHICH THE PREDICTION IS BEING MADE. BEST RESULTS IN EACH ROW ARE IN BOLD FONT.

| Future Timesteps ($k$) | RNN | | LSTM | | GRU | | Transformer | |
|---|---|---|---|---|---|---|---|---|
|  | MAE | RMSE | MAE | RMSE | MAE | RMSE | MAE | RMSE |
| 1 hour | 15.183 | 25.262 | 14.997 | 23.804 | **14.107** | 23.782 | 14.882 | **23.573** |
| 2 hours | 18.332 | 29.846 | 16.983 | 28.038 | 16.555 | 28.479 | **15.126** | **27.421** |
| 4 hours | 31.687 | 47.468 | 29.880 | 44.468 | 30.113 | 45.321 | **23.459** | **38.115** |
| 8 hours | 44.599 | 66.166 | 42.964 | **64.626** | 42.923 | 65.910 | **41.854** | 64.641 |
| 16 hours | 53.424 | 75.301 | 50.662 | 72.225 | 52.950 | 73.053 | **50.339** | **72.218** |

TABLE IV
PERFORMANCE (MAE AND RSME) OF SINGLE-STEP PREDICTION (LOOK-BACK WINDOW (4 DAYS) TO PREDICT 1 HOURS AHEAD) SHOWN AS A FUNCTION OF $w$, THE SIZE OF THE LOOK-BACK WINDOW USED FOR THE PREDICTION. BEST RESULTS IN EACH ROW ARE IN BOLD FONT.

| Look-back Window | RNN | | LSTM | | GRU | | Transformer | |
|---|---|---|---|---|---|---|---|---|
| | MAE | RMSE | MAE | RMSE | MAE | RMSE | MAE | RMSE |
| 1 day | 13.895 | 24.109 | 12.893 | **23.492** | **12.861** | 23.745 | 13.538 | 23.950 |
| 2 days | 14.878 | 24.958 | 13.566 | 24.090 | **13.180** | **23.878** | 13.946 | 23.896 |
| 4 days | 15.183 | 25.262 | 14.997 | 23.804 | 14.981 | 23.782 | **14.599** | **23.573** |
| 8 days | 17.031 | 26.998 | 16.089 | 24.379 | 16.892 | 24.886 | **15.255** | **23.499** |
| 16 days | 20.538 | 30.511 | 18.991 | 27.924 | 17.811 | 26.892 | **15.837** | **24.637** |

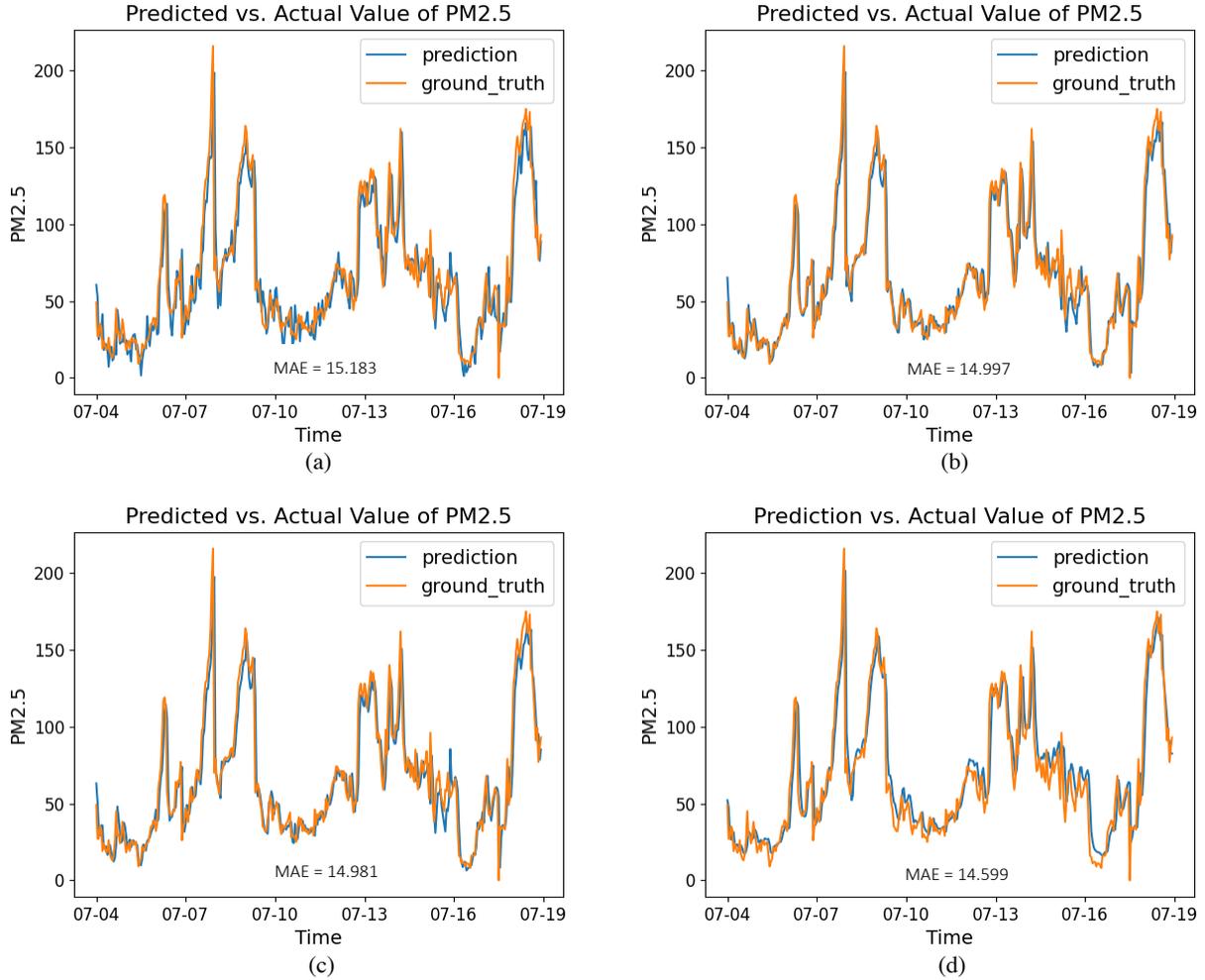

Fig. 8 Comparison of Predicted & Actual Values of PM2.5 (with $w = 96$ hours and $k = 1$ hour) using (a) RNN, (b) LSRM, (c) GRU, (d) Transformer models.

TABLE V
PERFORMANCE (MAE AND RSME) OF MULTI-STEPS PREDICTION (LOOK-BACK WINDOW (4 DAYS) TO PREDICT 3 HOURS AHEAD) SHOWN AS A FUNCTION OF $w$, THE SIZE OF THE LOOK-BACK WINDOW USED FOR THE PREDICTION. BEST RESULTS IN EACH ROW ARE IN BOLD FONT.

| Look-back Window | RNN | | LSTM | | GRU | | Transformer | |
|---|---|---|---|---|---|---|---|---|
| | MAE | RMSE | MAE | RMSE | MAE | RMSE | MAE | RMSE |
| 1 day | 27.732 | 43.645 | 26.225 | 42.564 | 25.360 | 40.625 | **23.998** | **38.767** |
| 2 days | 26.068 | 42.008 | 25.976 | 42.091 | 25.107 | 39.266 | **23.676** | **38.131** |
| 4 days | 27.729 | 43.429 | 25.994 | 41.872 | 25.818 | 39.501 | **23.273** | **39.889** |
| 8 days | 30.002 | 47.191 | 26.272 | 42.685 | 25.340 | 40.519 | **24.221** | **39.443** |
| 16 days | 35.958 | 53.742 | 29.991 | 45.816 | 28.887 | 44.284 | **26.192** | **43.889** |

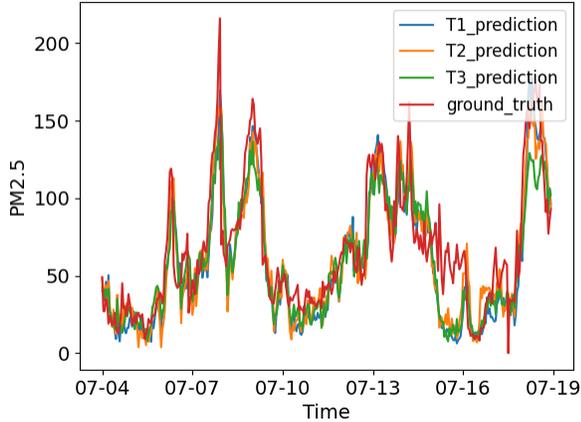
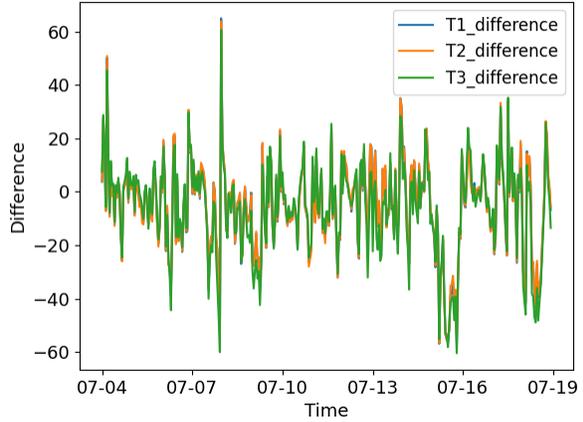

(a₁)                                (a₂)

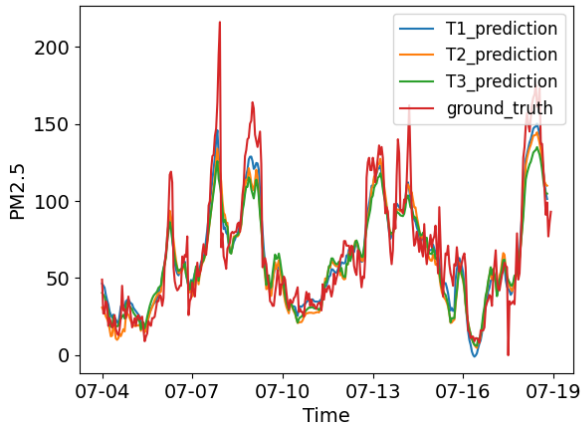
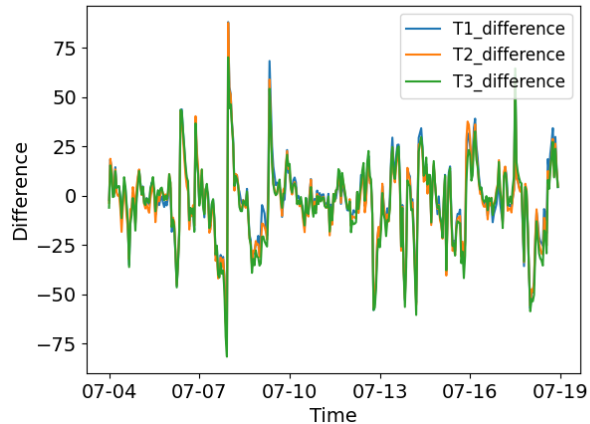

(b₁)                                (b₂)

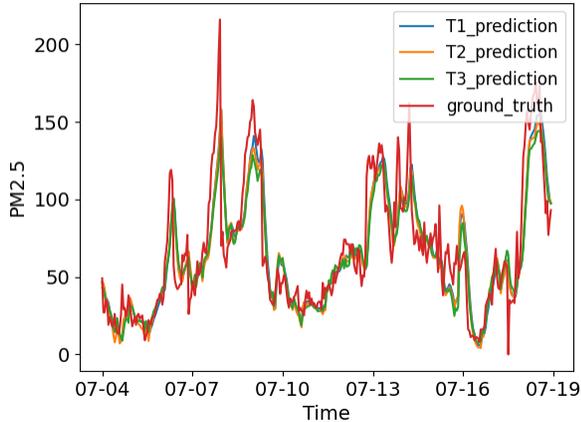
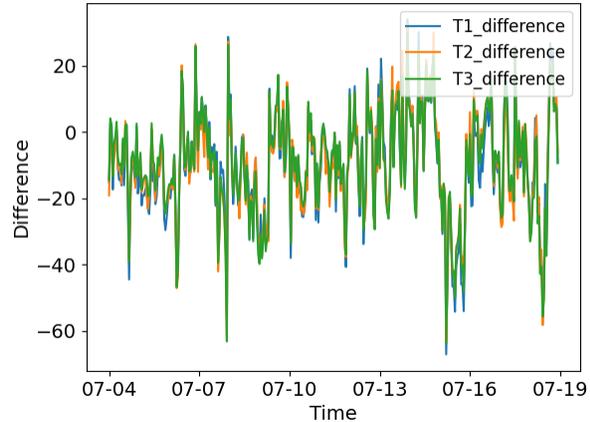

(c₁)                                (c₂)

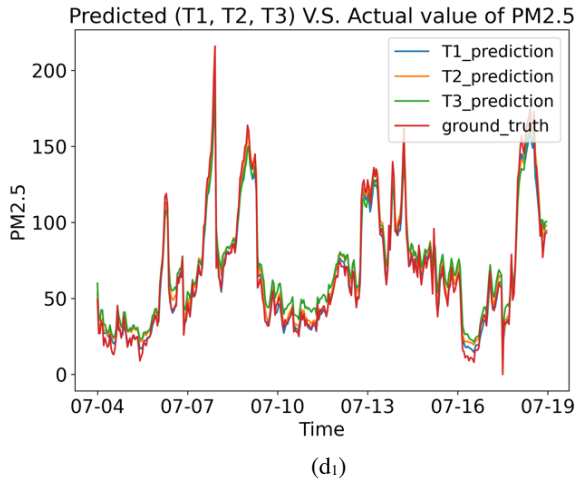 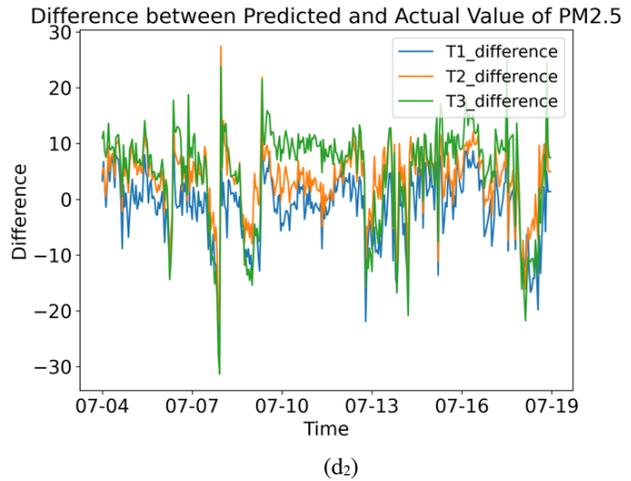

(d₁)     (d₂)

Fig. 9 Comparison of the predicted & actual values of PM2.5 (with $w = 96$ hours and $k = 3$ hours) using (a₁) RNN, (b₁) LSRM, (c₁) GRU, and (d₁) Transformer models. The plots on the right side (marked a₂, b₂, c₂, d₂) show the errors, which is the difference between actual and predicted values.

## VI. CONCLUSIONS

The conclusions from our experiments with four different deep learning models can be summarized as follows:

- Transformer network models perform the best when predicting farther into the future. LSTMs and GRUs outperform RNNs for shorter-term predictions.
- For the dependence of the performance on the look-back window size, the error displays a local minimum.
- For single-step predictions, the optimum value of window size is $w = 24$ hours. For multi-step predictions, the optimum value is $w = 48 \; or \; 96$ hours (when predicting $k = 3$ hours ahead).
- For multi-step predictions, Transformers outperform the other methods. For single-step predictions, Transformers perform well only if the look-back window is longer; GRUs and LSTMs are better for smaller $w$ values.

## APPENDIX

To understand the transformation process easier from original time series dataset to a supervised dataset, a simple example is given as below. We assume prediction timesteps forward are 1 and 2 for single-step and multi-steps respectively. Every prediction is based on a look-back window of length 3.

**Single-step Predictions:**

| Timestamp | x(t) | ŷ(t+1) |
|---|---|---|
| 1 | x(1) | ŷ(4) |
| 2 | x(2) | ŷ(5) |
| 3 | x(3) | ŷ(6) |
| 4 | x(4) | ŷ(7) |
| 5 | x(5) | ŷ(8) |
| 6 | x(6) | ŷ(9) |
| 7 | x(7) | … |
| 8 | x(8) | … |
| … | … | … |

**Multi-step Predictions:**

| Timestamp | x(t) | ŷ(t+3) |
|---|---|---|
| 1 | x(1) | ŷ(6) |
| 2 | x(2) | ŷ(7) |
| 3 | x(3) | ŷ(8) |
| 4 | x(4) | ŷ(9) |
| 5 | x(5) | ŷ(10) |
| 6 | x(6) | ŷ(11) |
| 7 | x(7) | … |
| 8 | x(8) | … |
| … | … | … |

ACKNOWLEDGMENT

We thank Vitalii Stebliankin for valuable suggestions and feedback on the work.